\documentclass{llncs}
\usepackage{llncsdoc}
\usepackage{epsfig,subfigure, float}

\begin{document}

\title{Swarmrobot.org -- Open-hardware Microrobotic Project for Large-scale Artificial Swarms}

\author{Serge Kernbach}

\institute{Institute of Parallel and Distributed High-Performance Systems,\\
University of Stuttgart, Universitaetsstr. 38, 70569 Stuttgart, Germany,\\
Serge.Kernbach@ipvs.uni-stuttgart.de}
\maketitle

\begin{abstract}
The purpose of this paper is to give an overview of the open-hardware microrobotic project swarmrobot.org and the platform Jasmine for building large-scale artificial swarms. The project targets an open development of cost-effective hardware and software for a quick implementation of swarm behavior with real robots. Detailed instructions for making the robot, open-source simulator, software libraries and multiple publications about performed experiments are ready for download and intend to facilitate exploration of collective and emergent phenomena, guided self-organization and swarm robotics in experimental way.
\end{abstract}

\section{Introduction}

The high miniaturization degree of the robotic systems with increasing efficiency of the hardware and software represents an important trend in the area of collective systems~\cite{Kernbach11-HCR}. The subarea -- swarm robotics -- is an emerged research field focused on designing collective intelligent systems comprised by a large number of robots. Its fascination and theoretical foundations originate from studying and understanding a group behavior of animals and insects societies~\cite{Bonabeau99}. It is expected that in a similar way a group of relatively simple and cheap robots can solve complex tasks that are beyond the capabilities of a single robot~\cite{Martinoli99}.

Natural physical and biological systems, being a metaphor for the phenomenon of self-organization, can organize themselves to emerge complex macroscopic behavior or spatiotemporal ordered structures and formations~\cite{Haken83}. There is no central element coordinating the system, a group behavior (macroscopic level) emerges from individual set of local rules (microscopic level) via interactions and cooperation. However, natural systems have multiple capabilities of interactions, perception and communication, which are significantly more complex than those in technical systems. Microrobots, due to a small size, are very restricted in locomotion, sensing and communication. Therefore a swarm-like behavior, expected from microrobotic systems, can "approach" a complexity of natural phenomena only when the robots are specifically developed, equipped and highly optimized for the desired collective activities~\cite{Kornienko_S04a}.

\begin{figure}[ht]
\centering
\includegraphics[width=1.\textwidth]{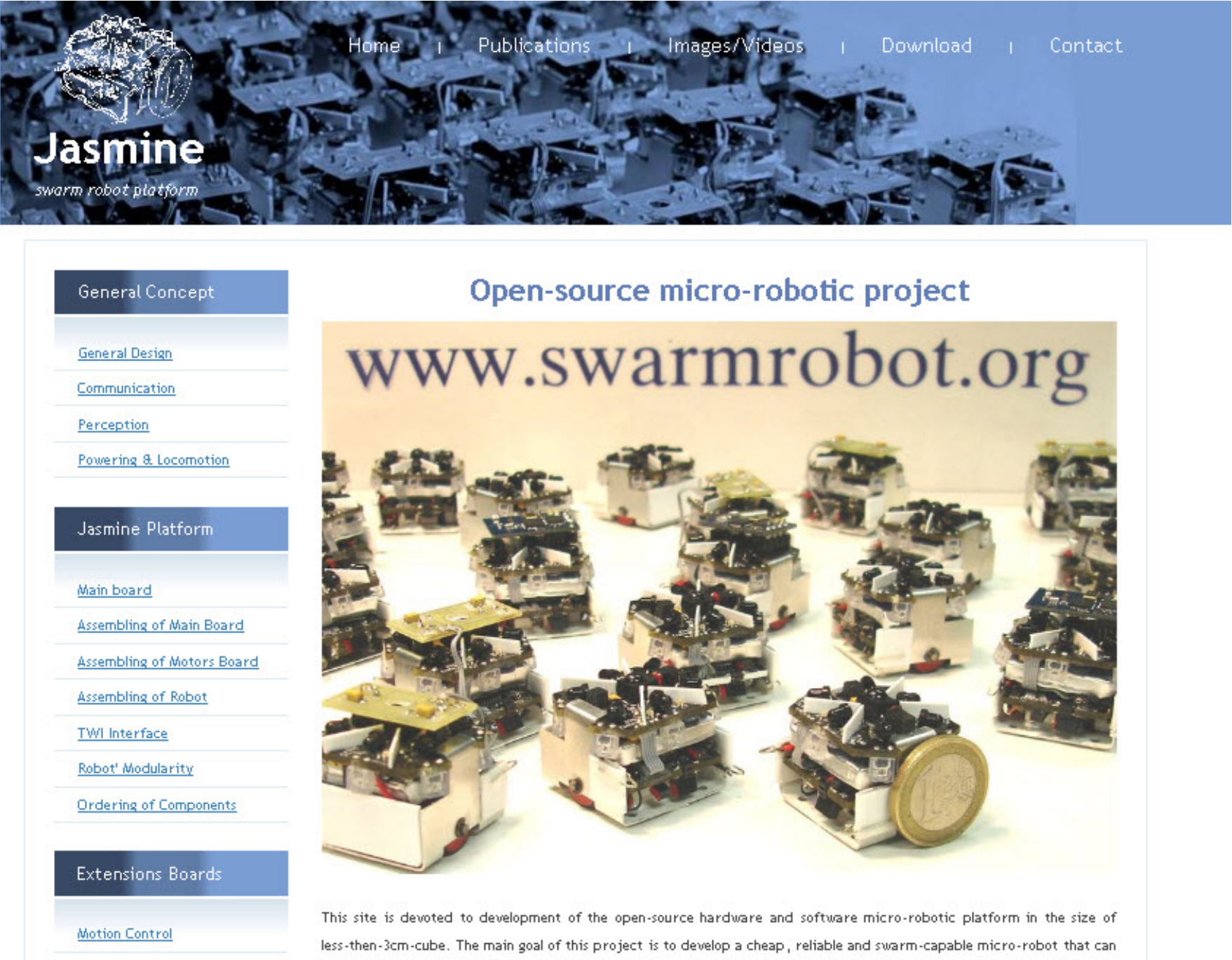}
\caption{\small \emph{Swarmrobot.org}, which is devoted to the platform Jasmine and large-scale swarm experiments.
\label{fig:swarmrobot} }
\end{figure}

Being motivated by the challenge of creating artificial self-organizing and emergent phenomena, we started open-hardware microrobotic research project for developing a specialized robot platform for large-scale swarms. It is based on the concept of swarm embodiment and is completely open at \emph{www.swarmrobot.org}, see Fig.~\ref{fig:swarmrobot}, for contributions from the domains of electrical and mechanical engineering, computer science and biology.
The project consists of several parts: a development of cheap and reliable hardware platform, an integrated SDK, open-source simulator and a description of the preformed experiments. All of them should allow a quick implementation of swarm behavior with real robots, exploration of collective phenomena emerged by artificial self-organization and may be of interest for broad areas of research and education. Since the project is maintained by universities, we can neither manufacture the robots for other participants nor perform any commercial activities with the platform -- the project swarmrobot.org targets primarily scientific goals.

This paper is structured in the following way: Section \ref{sec_embodiment} briefly describes the concept of swarm embodiment, which underlies the development. Sections \ref{sec_hardware}, \ref{sec_software} and \ref{sec:simulator} survey hardware, software and simulator for the robot platform. Finally, in Section \ref{sec_experiments} we overview performed experiments with Jasmine robots.

\section{Concept of swarm embodiment}
\label{sec_embodiment}

Swarm embodiment represents one of the main motivations for starting an open-hardware microrobotic development. The origin of this concept lies in the phenomenon of artificial self-organization (SO). Artificial SO differs from natural one in a couple of essential points, where a purposeful character of SO (so-called guided self-organization) represents the most important aspect. It means that a developer creates artificial SO to achieve some desired emergent effects, such as a specific group behavior of robots or desired molecular self-assembled structures.

Artificial SO is created by local rules. Each participant, driven by these local rules, demonstrates an individual behavior. All individual behaviors are macroscopically observable in the form of a coherent collective behavior. From the viewpoint of an observer, the collective behavior is emerged from interactions among swarm agents, which are guided by local rules. Derivation of local rules represent a serious problem, which is analytically non-solvable as proved by Poincare for the physical case of N-bodies interaction. There are three main strategies to derive (to approximate) such rules in non-analytical way. At the \emph{bottom-up strategy}, the local rules are first programmed into each agent~\cite{Roma93}. Performing many simulations and gradually changing the local rules, a desired collective behavior can be derived. Using the \emph{top-down strategy}, the derivation of local rules starts from a definition of the macroscopic pattern and the corresponding constraints. Then, using "distributing" transformation or evolutionary strategies, this macroscopic pattern can be transformed into a set of local rules, that in turn generate the desired pattern~\cite{Kornienko_S04b}. The last way consists in \emph{observing the behavior} of already existed artificial and natural swarm-like systems and in trying to reproduce this behavior~\cite{Bonabeau99}. All these ways lead finally to the local rules that control the behavior of each swarm agent.

The researcher, once derived these local rules and intended to implement them in a real robotic or in a simulation system, can face the questions "\emph{which general degree of collective intelligence is feasible in the destined system?}" and "\emph{how to implement it with the obtained local rules?}" To illustrate this point, we collect in Table~\ref{tab_colactivities} several swarm activities that robots can collectively perform.
\begin{table}[ht]
\centering
\caption{\small Several types of collective activities performed by a swarm. \label{tab_colactivities}}
\begin{tabular}{c | l | l  } \hline
{\bf Context} & \textbf{N} &{\bf Swarm Capability}\\\hline
           &  1   & Spatial orientation            \\
Spatial    &  2   & Building spatial structures    \\
           &  3   & Collective movement            \\ \hline
           &  4   & Building informational structures  \\
           &  5   & Collective decision making     \\
Information&  6   & Collective information processing \\
           &  7   & Collective perception/recognition \\ \hline
           &  8   & Building functional structures  \\
           &  9   & Collective task decomposition   \\
Functional &  10  & Collective planning             \\
           &  11  & Group-based specialization      \\ \hline\noalign{\smallskip}
\end{tabular}
\end{table}
We can roughly say, that these collective activities represent some building blocks for purposeful design of the swarm intelligence. Let us take the most simple example of spatial orientation and assume a robot has found a "food source" being
relevant for the whole swarm. This robot, guided by local rules, sends the following message: "I, robot X, found Y, come to me". Other robots, receiving this message can propagate it further through a swarm, so that finally each robot knows "there is a resource Y at the robot X". However, none of robots can find it because they do not know a coordinate of this "food source". The robot X cannot provide these coordinates because it does not know its own position. In this way, even possessing corresponding local rules and capabilities to communicate, robot cannot execute the desired collective activity "find food source" without some additional efforts intended to localization. Considering the localization problem, we face in turn the next generation of questions "\emph{how to implement it?}" In this way, {\it the original problem of local rules, generating a desired self-organization, changes into the problem of their implementation in real systems.}

More generally, during a derivation of local rules $R_k$ we assume some basic functionality $F_b$, like message transmission, localization or environmental sensing. Often we do not take into account real restrictions underlying this basic functionality or they are generally unknown at this step. However, implementing later these $R_k$, we obtain the swarm behavior, that differs from our expectations (even with correctly derived $R_k$), because a real functionality $F_b$ can essentially differ from the expected one.

\begin{figure}[ht]
\centering
\includegraphics[width=.7\textwidth]{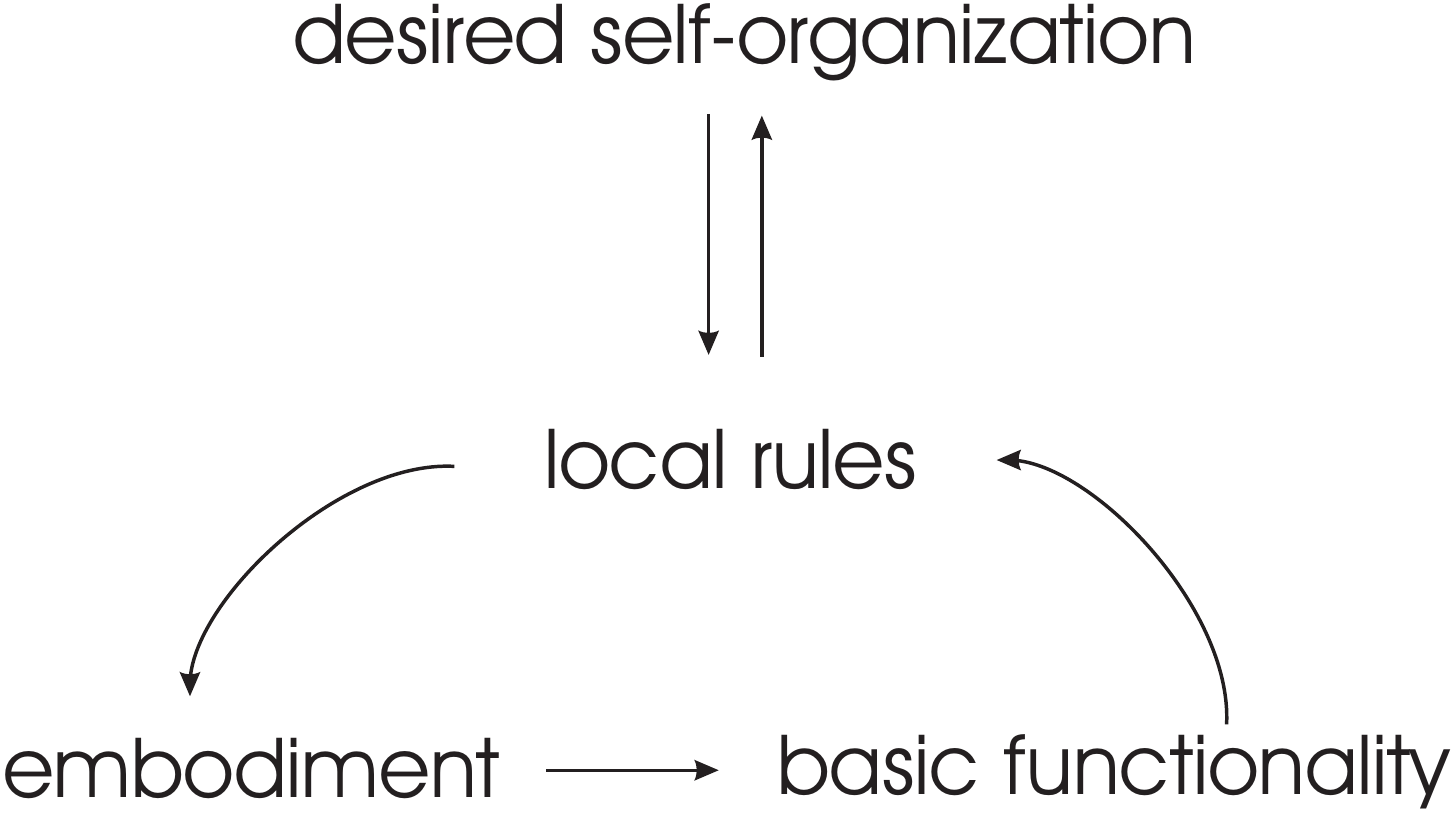}
\caption{\small Concept of embodiment of local rules. \label{fig_embody}}
\end{figure}

To get round this problem, we involved the embodiment concept (see Fig.~\ref{fig_embody}). This says the same functionality can be implemented in many different ways: "intelligent behavior" can even be implemented when using only some properties of materials~\cite{Pfeifer04}. Embodied functionally is also often implemented in some "unusual" way. For example a robot can estimate a distance to neighbors by sending an IR-impulse and measuring a reflected light. However, distances can also be obtaining during communication by measuring a signal intensity. This simple mechanism saves time and energy: such an unusual functionality is a typical sight of embodiment.

The embodiment in our context means the system possesses the desired functionality $F_b$, but this functionality is in a latent form, "it is not appeared". This offers a way of how to get basic functionality for the local rules $R_k$: the local rules have to influence the hardware development of a robot. The swarm embodiment takes then the following form: definition of the macroscopic pattern $\Omega$ and the corresponding microscopic/macroscopic constraints; derivation of the local rules $R_k$; trade off between required functionality and adjustment of
hardware; change of the hardware. The local rules have always been considered as a pure software components, however now they are a combination between software and hardware. We can say that in this way \emph{the local rules for the whole swarm behavior are embodied into each individual robot}.

The embodiment in sense "\emph{hardware} $\rightarrow$ \emph{rules}" has been demonstrated e.g. in works~\cite{Kornienko_S04b} \cite{Kornienko_S05e}, where we analyzed a dependence between agent's movement and sensor data for the derived $S_k$. Optimizing local rules to specific motion system, the ``top-down"-derived rules can be 5-20\% more efficient than corresponding ``bottom-up" rules. The embodiment in sense "\emph{rules} $\rightarrow$ \emph{hardware}" has been demonstrated e.g. in works \cite{Kornienko_S05b}, \cite{Caselles05}. In these works we considered context awareness related collective capabilities of interacting robotic group and incorporate several local rules into specific sensor system of real microrobots. The achieved results improve collective robotic behavior and reduce required communication and computational efforts.

After those experiments we came to conclusion that {\it swarm robot differs from a general-purpose robot}, even when this robot is small enough. The swarm-capable robot should not only incorporate in its own construction some specific details, such as capabilities of swarm communication or cooperative actuation, but be open for embodiment of different local rules. It should have some open-hardware structure. This was the main motivation to derive new microrobotic platform for large-scale
swarm experiment and make it in open-hardware manner.

\section{Open-hardware development}
\label{sec_hardware}

\begin{figure}[h!]
\centering
\includegraphics[width=.5\textwidth]{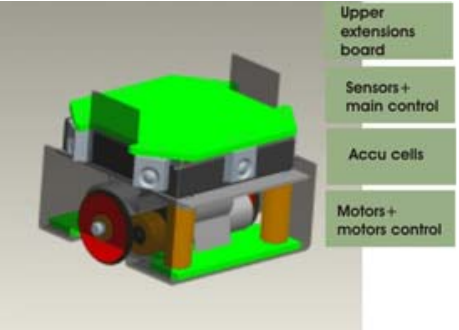}
\caption{\small "Sandwich design" of the Jasmine platform.
\label{fig_jasmine3} }
\end{figure}

The design of the robot, assembling instructions, electrical schemes and PCB can be found at \textbf{www.swarmrobot.org}. Here we briefly sketch several main points:
\begin{enumerate}
    \item
One of the most important requirements is that the microrobot platform for large-scale swarm experiments should be \textbf{cheap}. We require that the swarm robot should cost at most 100 euro. Obviously, this requirement imposes several compromises on the capabilities of the platform.

\item The robot should be small, no larger than a cube with the edge of 30 mm. This size hardly limits hardware, so that we come closely to real "limited swarm agent".

\item It should be easy in assembling so that each research institute or even private persons can produce enough robots for their own swarms. This means primarily avoiding SMD soldering and a high-precision micro-mechanical assembling.

\item The robot should have enough hardware resources (ROM, RAM, MPU frequency) to maintain at least a simple operational system with users-defined software.

\item Rich communication and perception capabilities oriented for large- scale swarms. This means: omni-directional communication system; communication radius, which is large enough for a reasonable information transfer in a swarm; support for context-based features; perception system, which allow recognition of main object features such as geometry or  colors.

\item The running time without a new recharging should be at least 1-2 hours for the hardest conditions (full speed motion, continuous communication and perception);

\item The robot has to be extendable (modular) for other sensors/actuators/ communication boards with different embodiment rules. This means: standardized inter-board logical, mechanical and electrical interfaces; standardized robots size and structural construction.
\end{enumerate}
General design of the microrobot is limited by the size and geometrical configurations of used components. Since we would like to have a robot that is small as possible, but still cheap and easy in assembling, the most efficient structure in this case is ether an integration of all components on one PCB (flexible PCB) or a "sandwich design" (see Fig.~\ref{fig_jasmine3}).
Analyzing sensors for sensing and communication we come at a conclusion that SMD sensors are not really suitable. Thus, an integration of all components into one PCB is not possible and so only "sandwich design" can be used. Order of "sandwich layers" is defined by the size of accumulator: it can be placed only between the motor PCB and sensors board. All layers of this "sandwich" can be covered by a thin plastic or, more sophisticated, by metal chassis. This design is simple for assembling, cheap and protect electronics and mechanics from possible damages.

\begin{figure}[htp]
\begin{center}
\subfigure[]{\includegraphics[width=.47\textwidth]{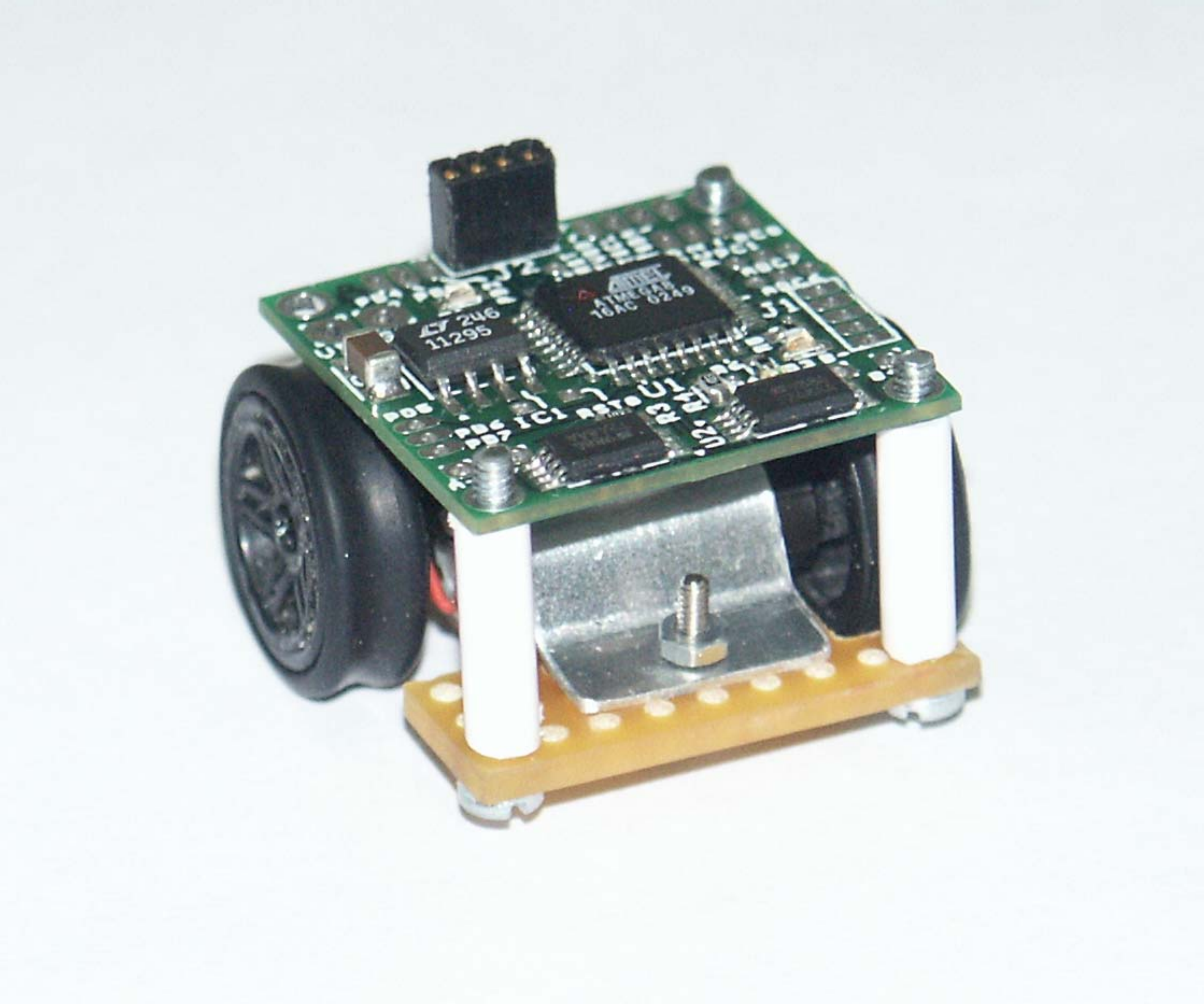}}~
\subfigure[]{\includegraphics[width=.49\textwidth]{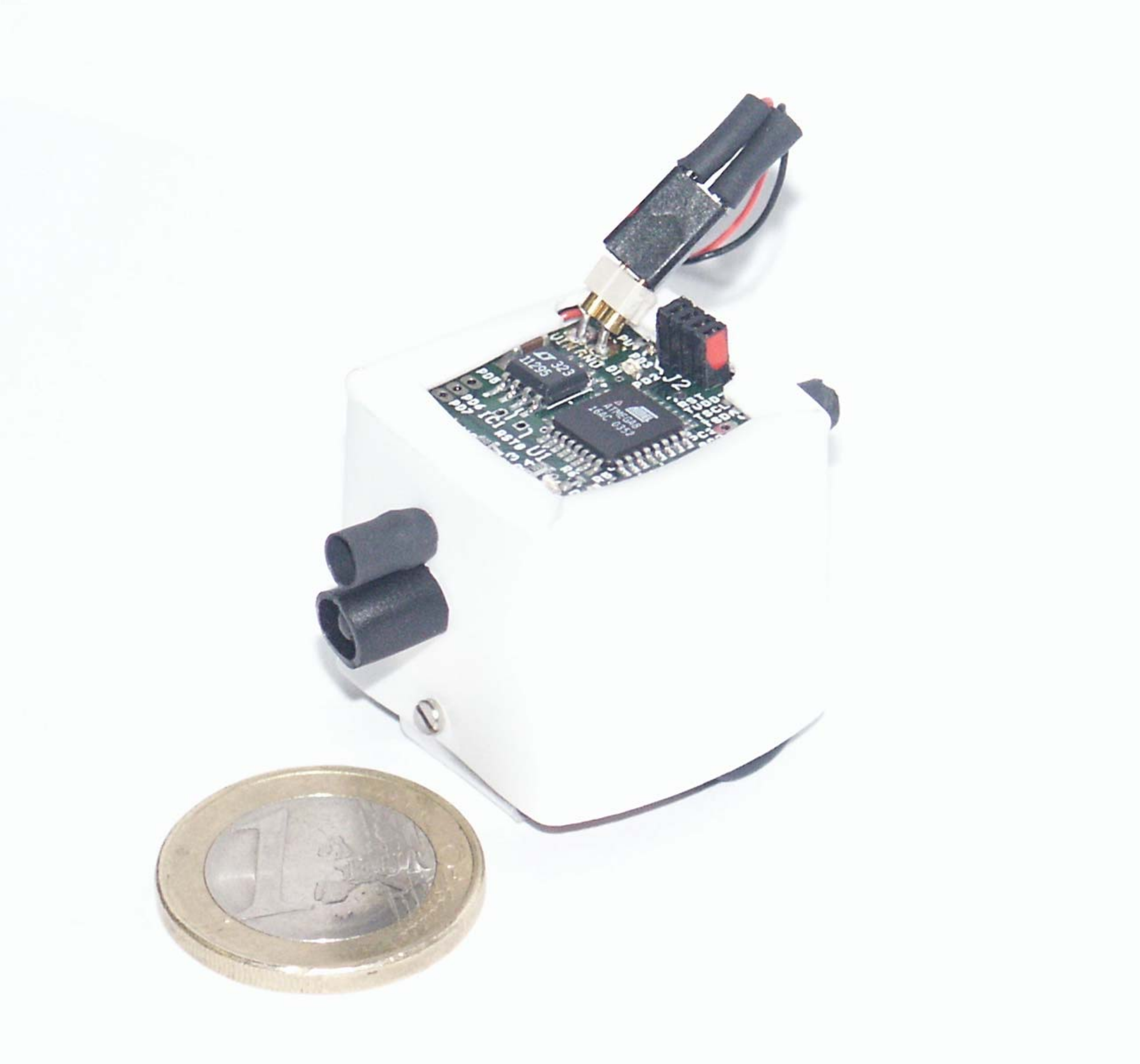}}\\
\subfigure[]{\includegraphics[width=.52\textwidth]{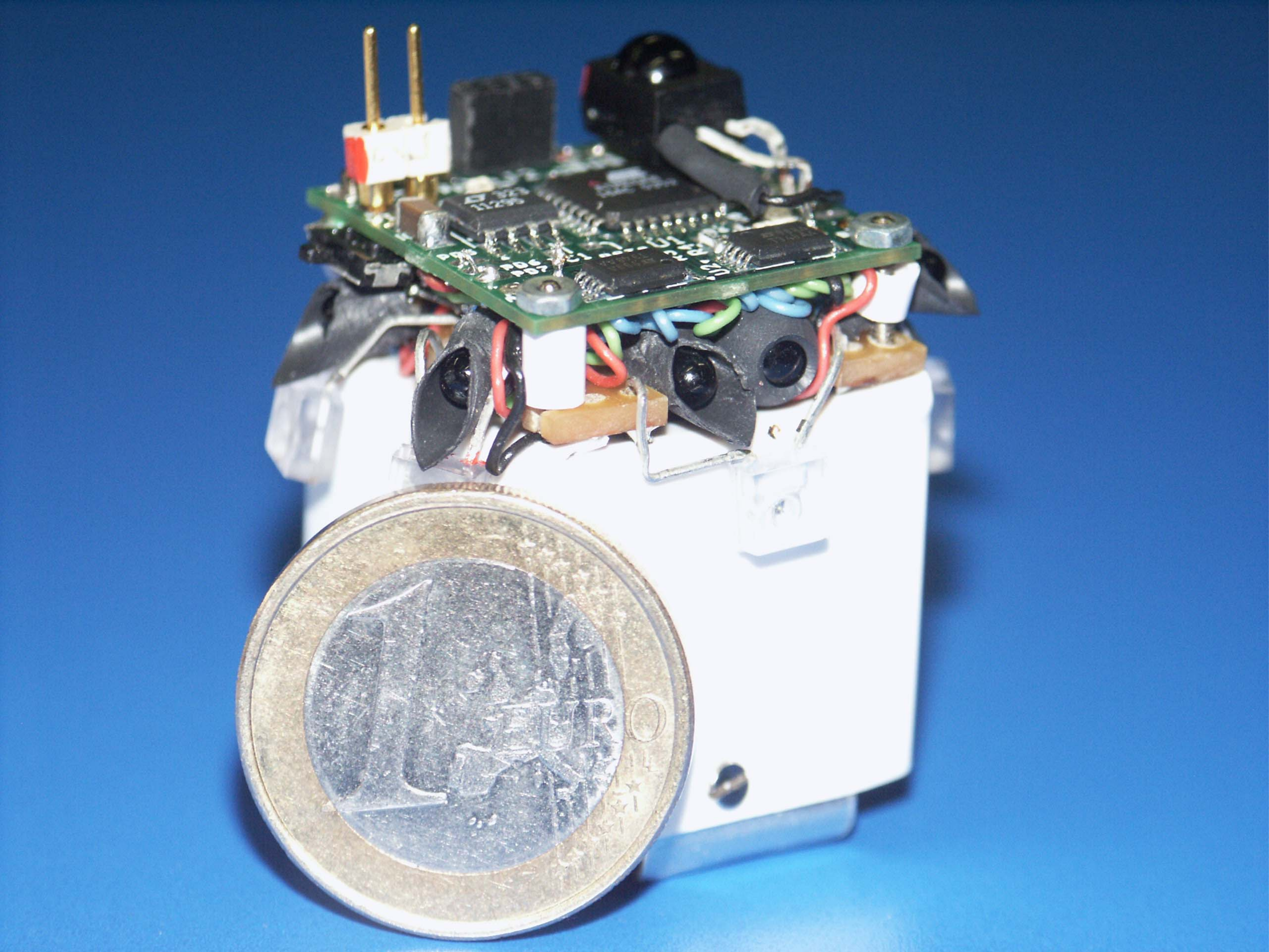}}~
\subfigure[]{\includegraphics[width=.475\textwidth]{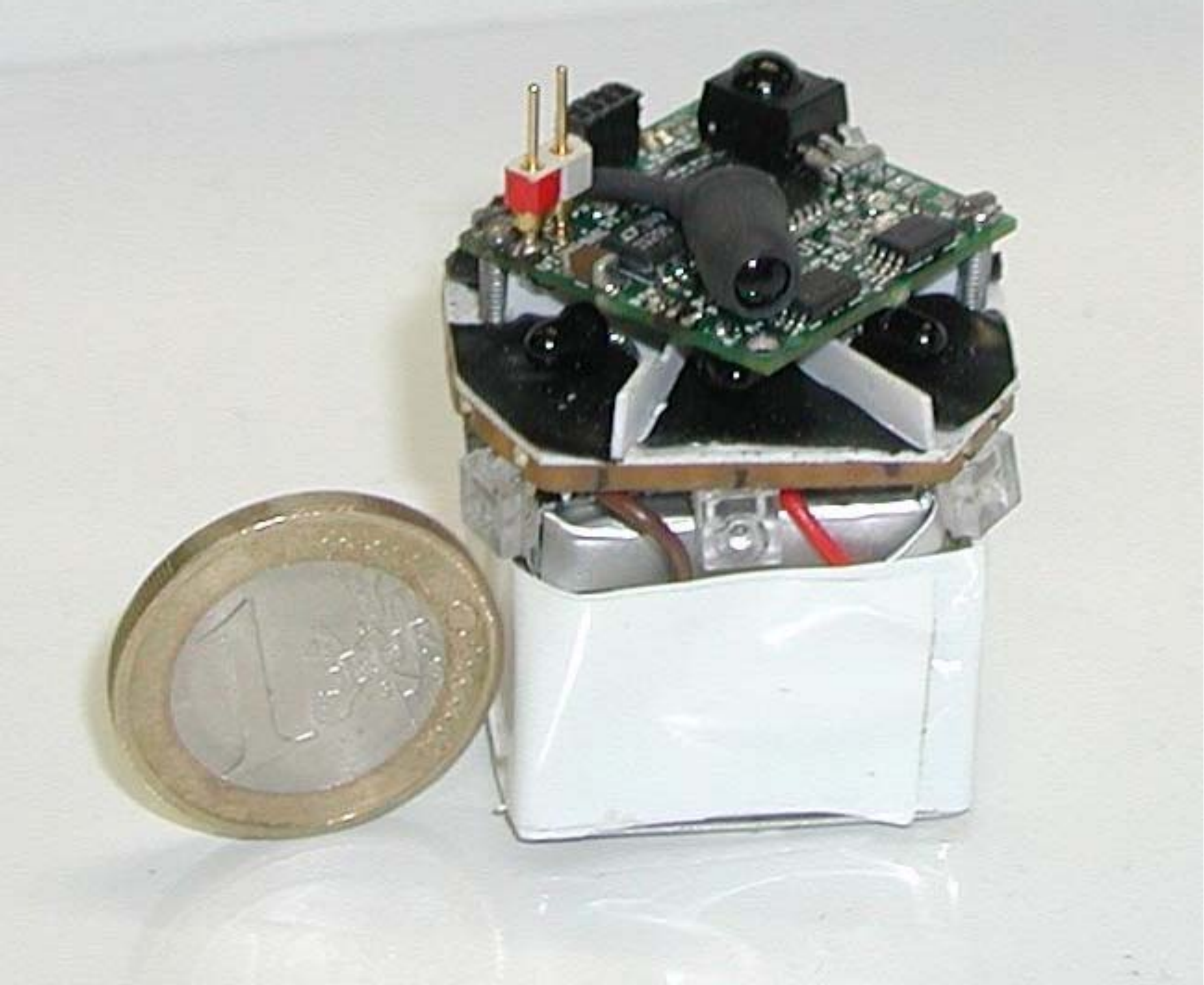}}\\
\subfigure[]{\includegraphics[width=.54\textwidth]{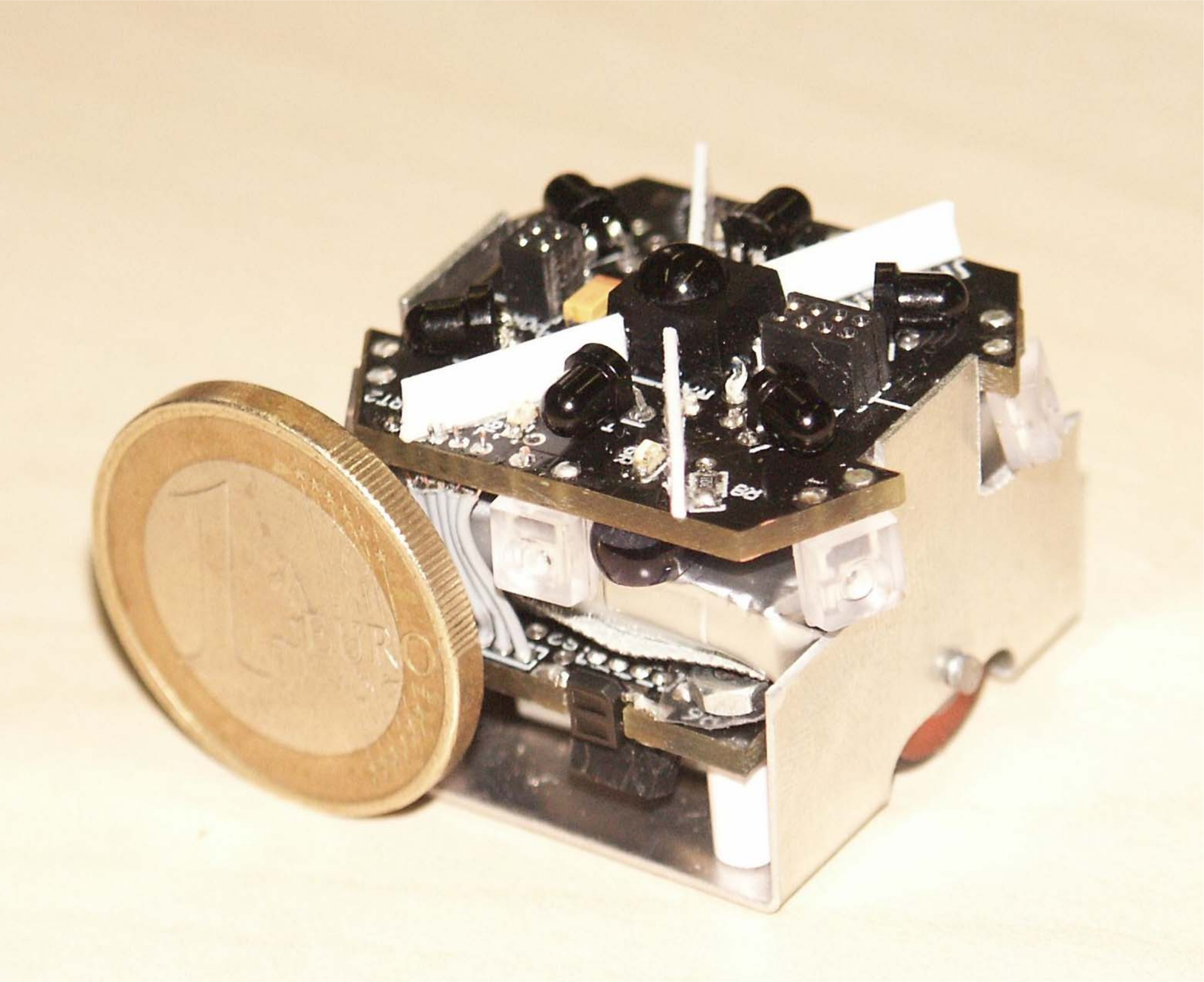}}~
\subfigure[]{\includegraphics[width=.46\textwidth]{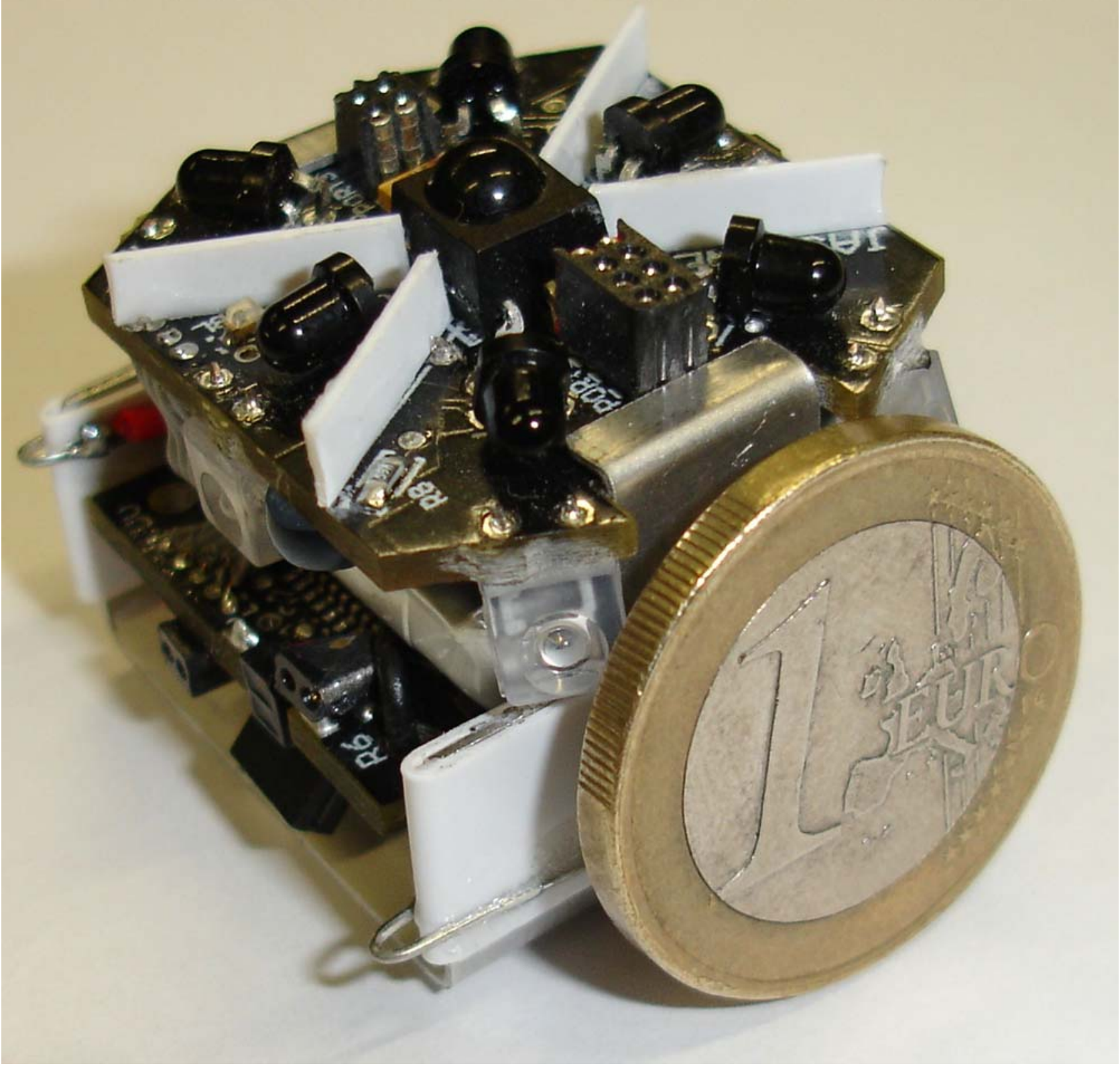}}
\caption{\small Different versions of the microrobot Jasmine.
{\bf (a)} Chassis with the megabitty board;
{\bf (b)} The first test development for testing of sensors;
{\bf (c)} The first version;
{\bf (d)} The second version;
{\bf (e)} The third version;
{\bf (f)} The third plus version;
\label{fig:Jasmine}
}
\end{center}
\end{figure}

The robot uses two GM15(RM-N1) motors with gear motor-wheel coupling, one Li-Po accumulator cell C=250mA/h with 3.7V, electronics of the robot works with 3V. Robot consumes about 200-220 mA in a full motion and communication modus, that is less that 1C and it has have enough capacity for running time of 1-2 hours. The sensing and communication capabilities of the Jasmine platform are already described (see e.g. for communication \cite{Kornienko_S05b} and for perception \cite{Kornienko_S05d}). Briefly, the robot can recognize the surface geometry (concave, convex, round, flat), color of surfaces (at least three main
colors), can sense obstacles in 360 degree around itself, in a distance over 150mm, internal energy level and so on. Communication is omni-directional, half-duplex, max. communication radius is over 150mm (300mm in 15 degree opening angle). In Fig.~\ref{fig:Jasmine} we show different developed versions of the platform.

\begin{figure}[ht]
\centering
\includegraphics[width=.7\textwidth]{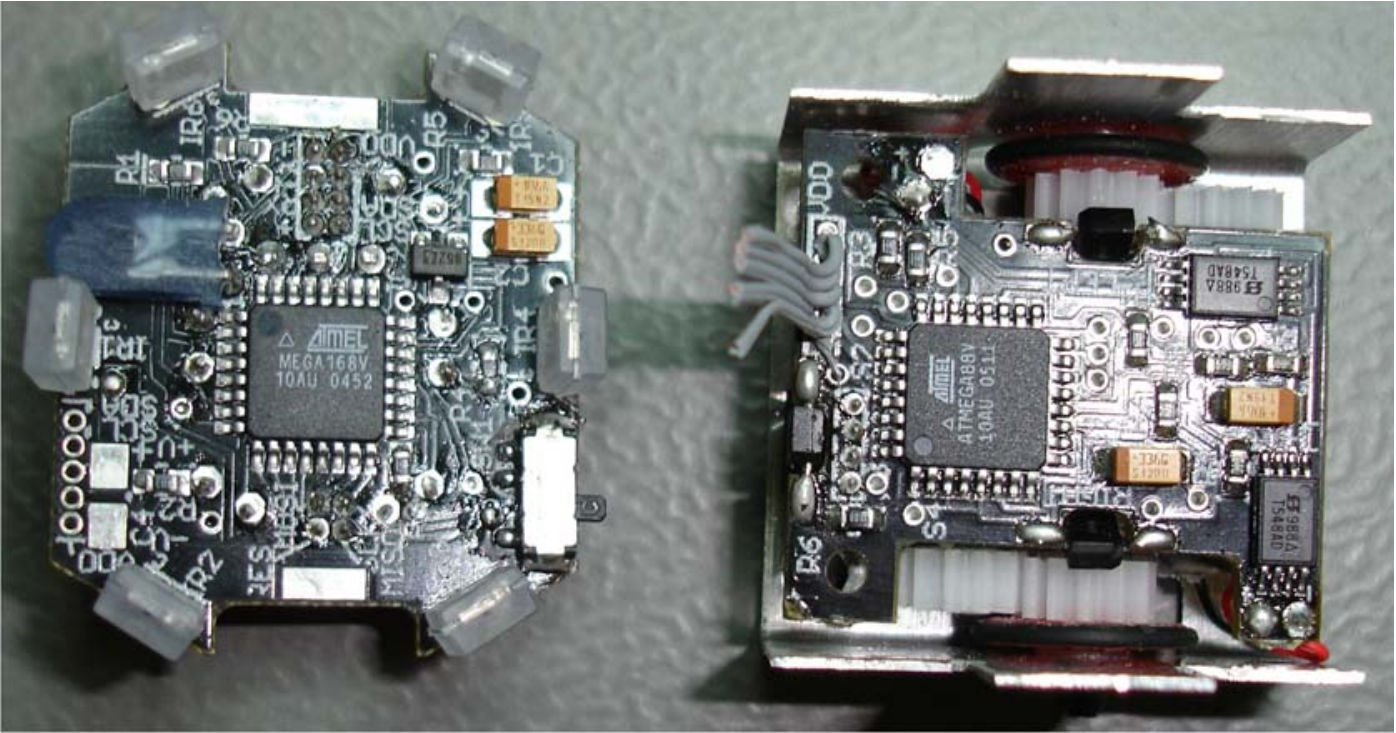}
\caption{\small Modularity of the Jasmine platform: the main board (left) and the motors board (right). \label{fig_modularity}}
\end{figure}

The microrobotic platform is modular, it means the robot can be configured as main board plus different extensions boards (see Fig.~\ref{fig_modularity}).
Currently there are a few extension systems, listed in the Table~\ref{tab_extensions} that are in different state of development/testing.
\begin{table}[ht]
\centering
\caption{\small Potential and already existed extension boards for Jasmine platform. \label{tab_extensions}}
\begin{tabular}{l | l   } \hline
{\bf Extension}      & {\bf  Description} \\\hline

Ego-position system  &  Allows determining the robot position/rotation  \\
                     & in global coordinate system with accuracy of a few mm.\\
                     & It requires a beamer on the top of robot-arena, \\
                     & software for beamer control and specific two-sensor \\
                     & system on the top of the robot.\\

Light sensors system & Allows following the light source.\\
                     & It requires two light sensor on the top of the robot.\\

Wireless communication   & Uses Atmel's RF chip on the top extension. \\
                         & It requires a relatively large antenna. \\
Odometrical system       & Allows determining the local robot displacement \\
                         & and rotation in with accuracy of a few mm. \\
                         & It required specific sensor system on the motor board. \\

Auto Recharge System  & Allows recharging of robot' accu during a motion. \\
                       & The robot can decide when its own accu needs to be \\
                       & recharged and does it autonomously without human\\
                       & participation. \\

Electro-magnetic gripper & Allows gripping small metallic objects (in development).\\
                       \hline
\end{tabular}
\end{table}

Finally, we summarize the main capabilities of the platform:
\begin{itemize}
\item low-price solution, as components less than 100 euro per robot;

\item size 26 x 26 x 26 mm;

\item RAM 2kB, Flash ROM 24kB,nonvolatile EEPROM 1kB;

\item 2 DC motors (reversible), max. velocity over 500 mm/sec.;

\item autonomous work 1-2 hours;

\item omni-directional 6 channel robot-robot communication, half-duplex;

\item physically adjustable communication radius: 0-max, max. communication over 300 mm, capacity of channel over 1000 bits/sec.;

\item remote control and host-robot communication;

\item proximity and distance measurement over 300 mm;

\item perception of objects geometries, light, tough, internal energy sensor and so on;

\item extendable periphery (e.g. camera) with serial interface, external actuators;

\item programmable via COM-Port from PC, many open-source development tools.
\end{itemize}

Since this is the open-hardware project, the hardware details can be downloaded from the site, otherwise in the community there are a few companies that can provide a complete solution with a hardware support.

\section{Control concepts and supported software}
\label{sec_software}

\begin{figure}[ht]
\centering
\includegraphics[width=.7\textwidth]{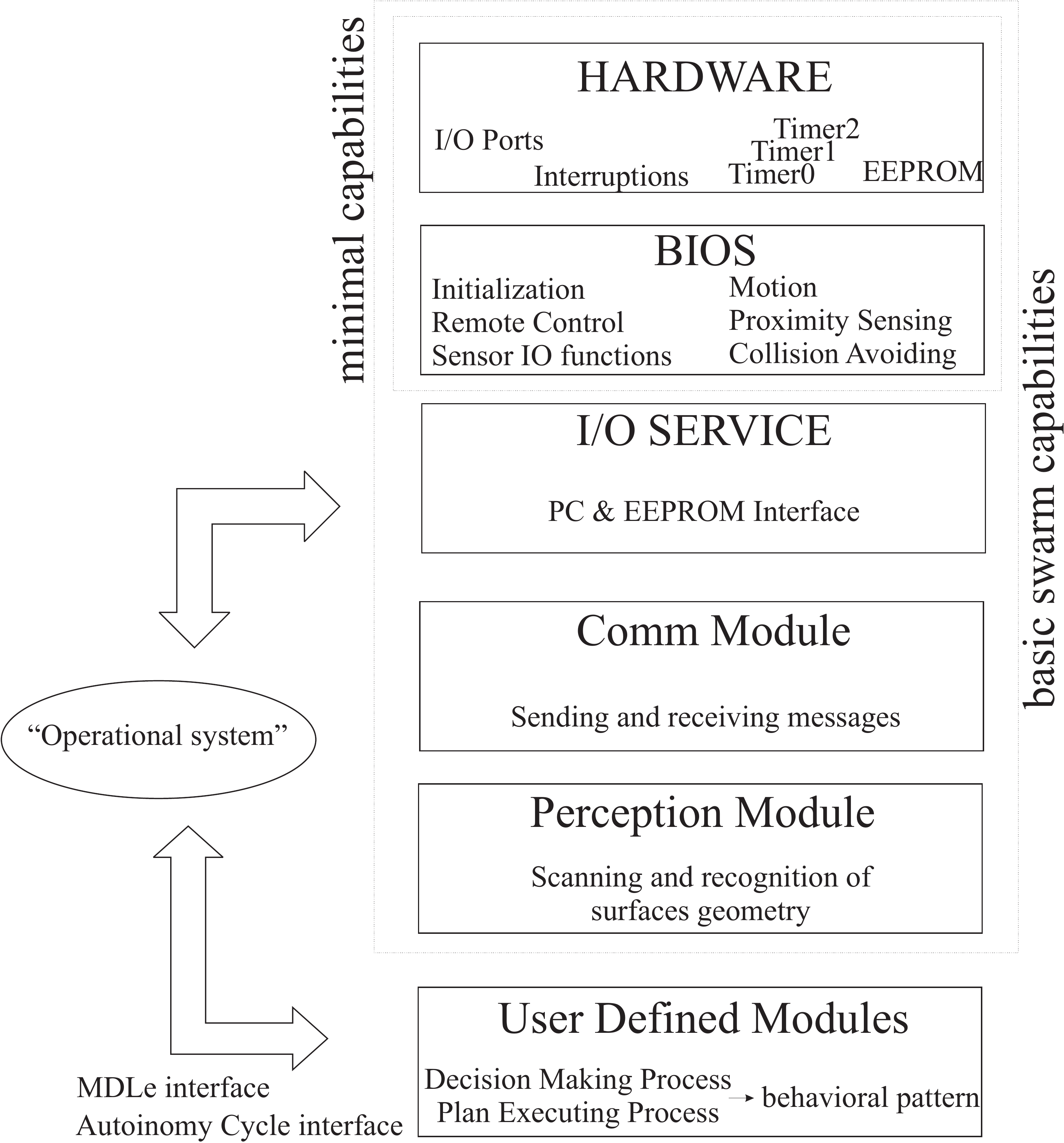}
\caption{\small Software structure of Jasmine platform.
\label{fig_software} }
\end{figure}

The software is structured hierarchically, where the high-level modules are based on the low-level modules, as shown in Fig.~\ref{fig_software}.
These high-level modules can be exchanged without changing basic capability of the robot. Since the high-level modules describe the behavioral scenarios, we can implement many different scenarios without overstepping hardware capabilities (flash memory) of the microrobot. The modular construction allows also a "distributed" software development by different persons, easy improvement and extension. Moreover, modules that are not required in a particular scenario, can be removed from a configuration without destroying the software architecture.

There are five main levels of the software architecture:

1. BIOS Module: This module provides the main interface to hardware resources. It supports remote control, motion control, proximity sensing, TWI interface, software interruption system, collision avoiding and so represents the basic capabilities of the
microrobot.

2. I/O Services: This module provide the interface to PC (serial interface) for reading/writting and writing into EEPROM.

3. Comm Module: Communication module provides a low-level half-duplex local data exchange between the robots. It supports two main protocol: with confirmation and without confirmation. The module is relatively large and can be removed from configuration (in scenarios where communication is not required). This module includes also a support for active communication (requires user-defined support).

4. Perception Module: Perception module is destined for scanning and recognition of surfaces geometry. This routine includes also basic functions for collective perception (requires user-defined modules).

5. User-defined module: This module includes primarily two functions \emph{DecisionProcess()} and \emph{ExecutePlan()} and determines the high-level behavioral patterns. Each user (working on separate scenario) creates its own user-defined module or uses them from share-library.

The "operating system" of the robot executes user-defined code and provides hardware resources, timing and other resources to a user-defined code. There are two operating systems that use either the autonomy cycle or software interruptions. Independently of the used operating system, the BIOS, perception and communication modules remain the same. These components build an integrated on-board SDK, which allows a quick and efficient implementation of swarm algorithms. The C-source and libraries for AVR compiler (open-source) can be downloaded from the site.

\section{Simulator}
\label{sec:simulator}

The simulator is used for preliminary tests and debugging the software before programming real robots.
\begin{figure}[h!]
\centering
\subfigure[\label{fig:sim1}]{\includegraphics[width=.51\textwidth]{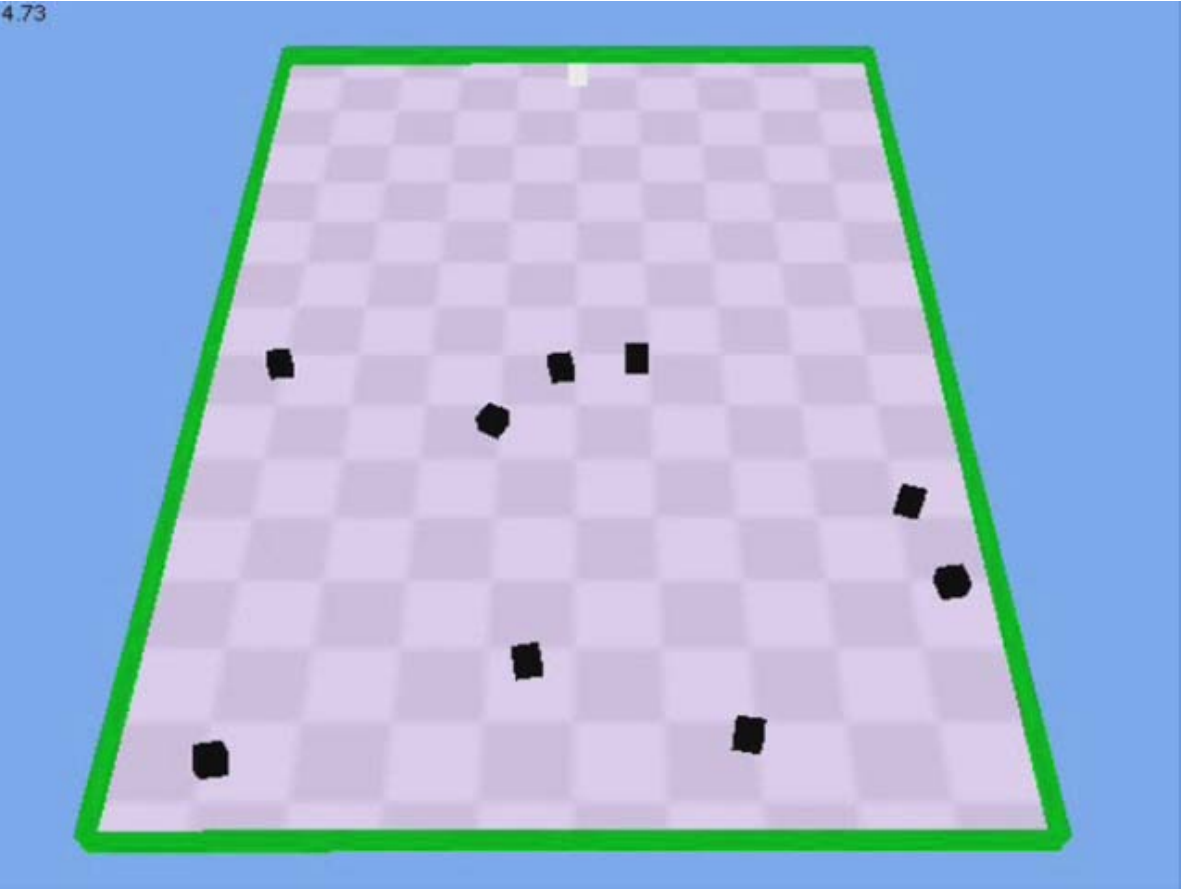}}~
\subfigure[\label{fig:sim2}]{\includegraphics[width=.48\textwidth]{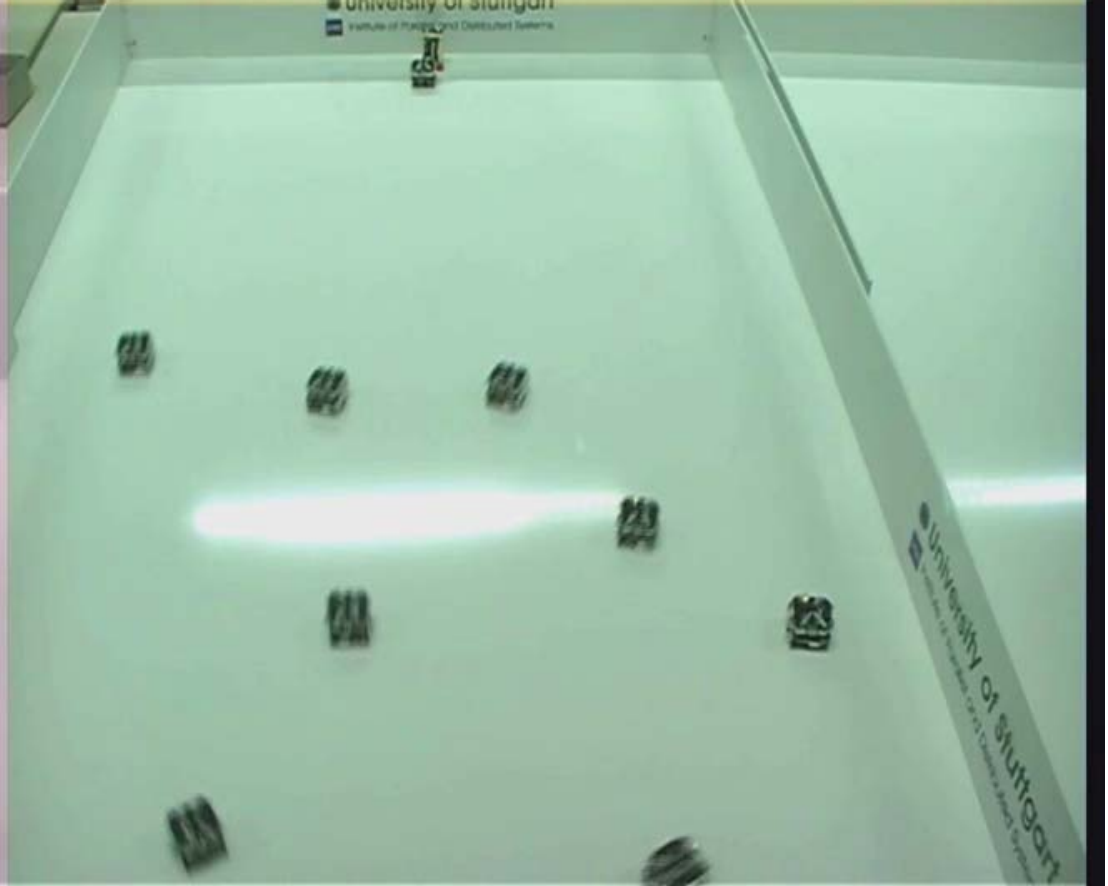}}\\
\subfigure[\label{fig:sim3}]{\includegraphics[width=.49\textwidth]{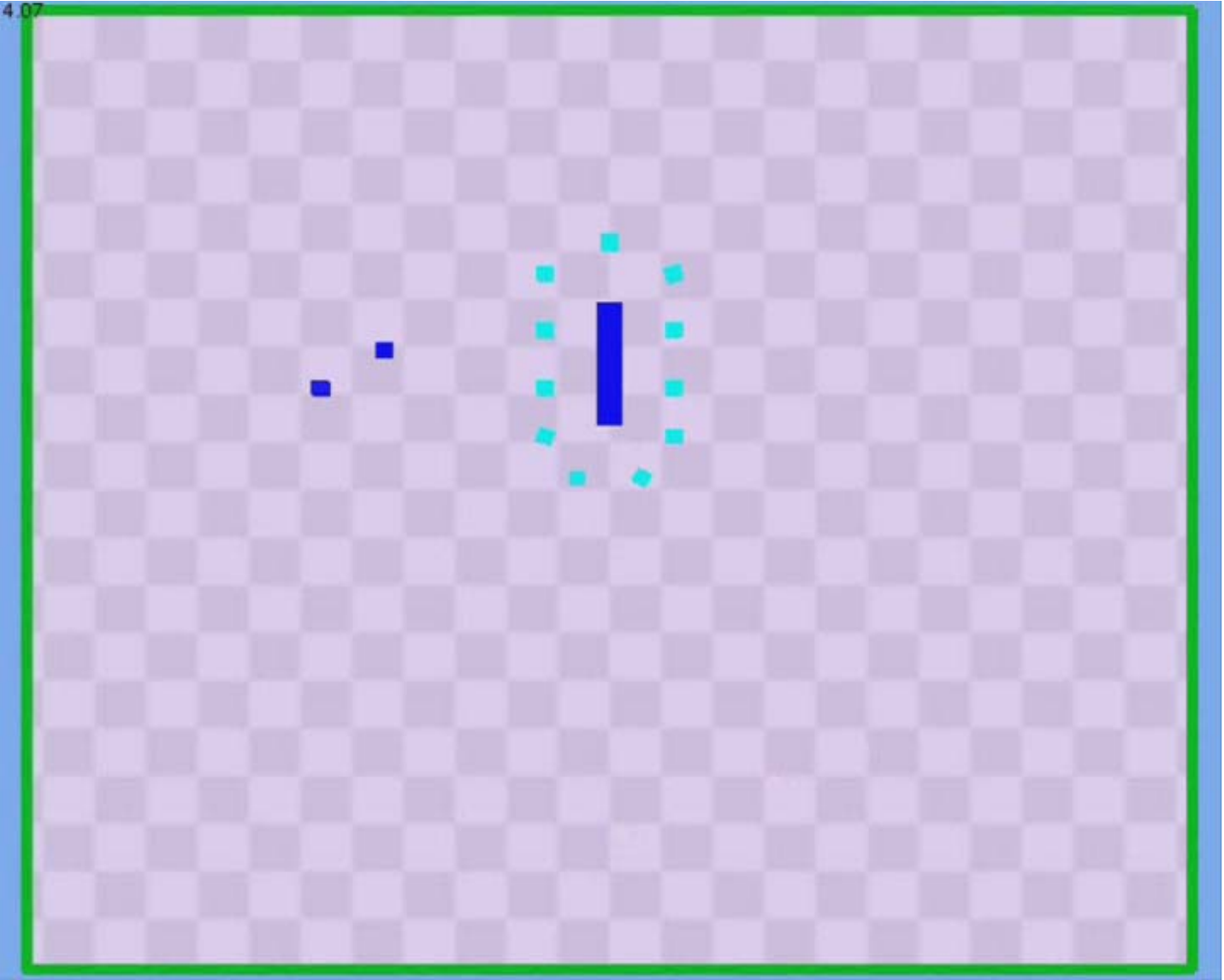}}~
\subfigure[\label{fig:sim4}]{\includegraphics[width=.49\textwidth]{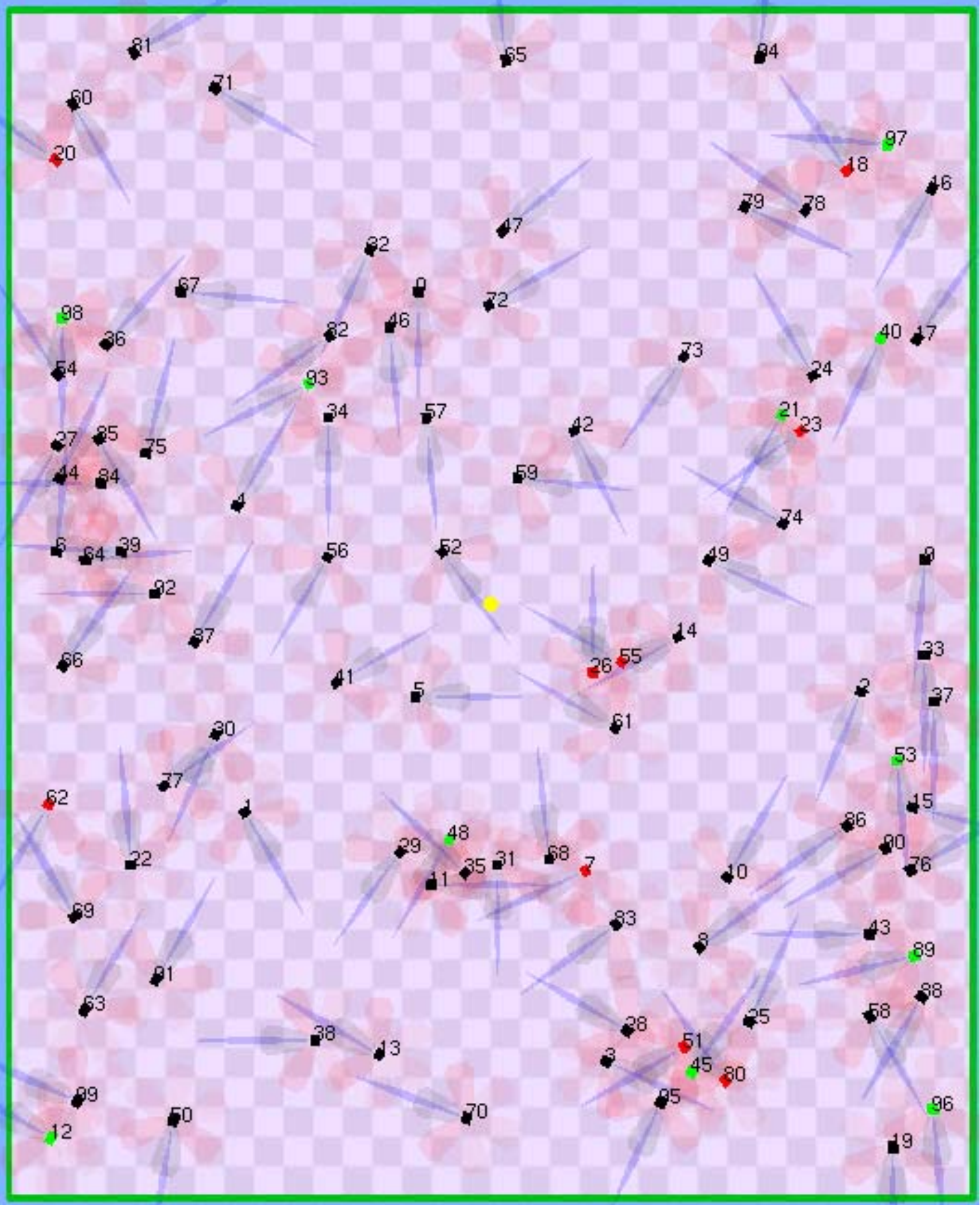}}
\caption{\small \textbf{(a)} Setup in simulation; \textbf{(b)} The same setup in real experiments; \textbf{(c)} Example of a setup for exploring collective perception; \textbf{(d)} Example of a setup for exploring large-scale swarm.
\label{fig:simulator} }
\end{figure}
The program used to create the simulation system is called 'Breve'. It is an open-source, multiplatform software package, which makes an fast development of 3D simulations for decentralized systems and artificial life. Users define the behaviours of agents in 3D world and observe how they interact. 'Breve' includes an OpenGL engine so it is possible to visualize simulated worlds. Breve simulations are written in an object-oriented language called Steve. More information about Breve can be found in \cite{Klein00}. Breve executes the code written in each simulation agent in a parallel way every iteration, so all agents are independent.

Especial attention was paid to create equal hardware abstraction layers in Jasmine robots and in the simulator so that the same code can be run in both systems without essential modifications. A large effort was invested into making reality gap as minimal as possible: for instance, we compared behavior between simulated and real setups, see Fig.~\ref{fig:sim1} and \ref{fig:sim2}, and tuned the simulator to obtain almost identical behaviors, see more in \cite{Prieto06}. Since all sensors are fully implemented, the simulator was employed even for preliminary tests with collective perception based on surface colors and surface geometries, see Fig.~\ref{fig:sim3}. Forasmuch as the reality gap between Jasmine robots and simulator became finally relatively low, it was also used for performing preparatory experiments with large-scale swarms, see Fig.~\ref{fig:sim4}, where the effort of performing real experiments with many attempts was especially high.

\section{Experiments with microrobots Jasmine}
\label{sec_experiments}

\begin{figure}[htp]
\centering
\subfigure[\label{fig:experiments1}]{\includegraphics[width=.48\textwidth]{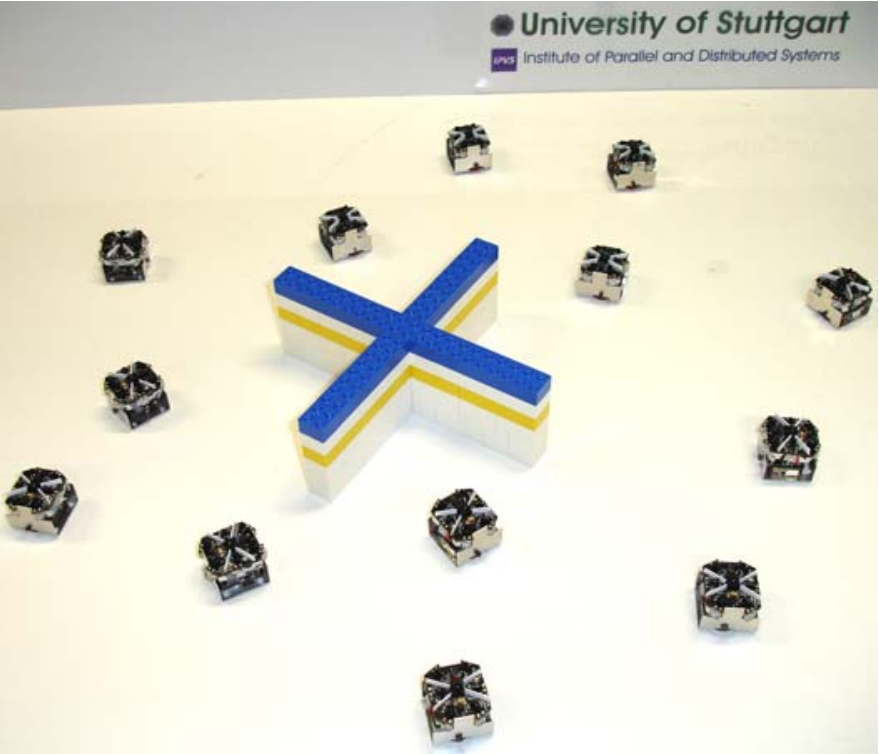}}~
\subfigure[\label{fig:experiments2}]{\includegraphics[width=.515\textwidth]{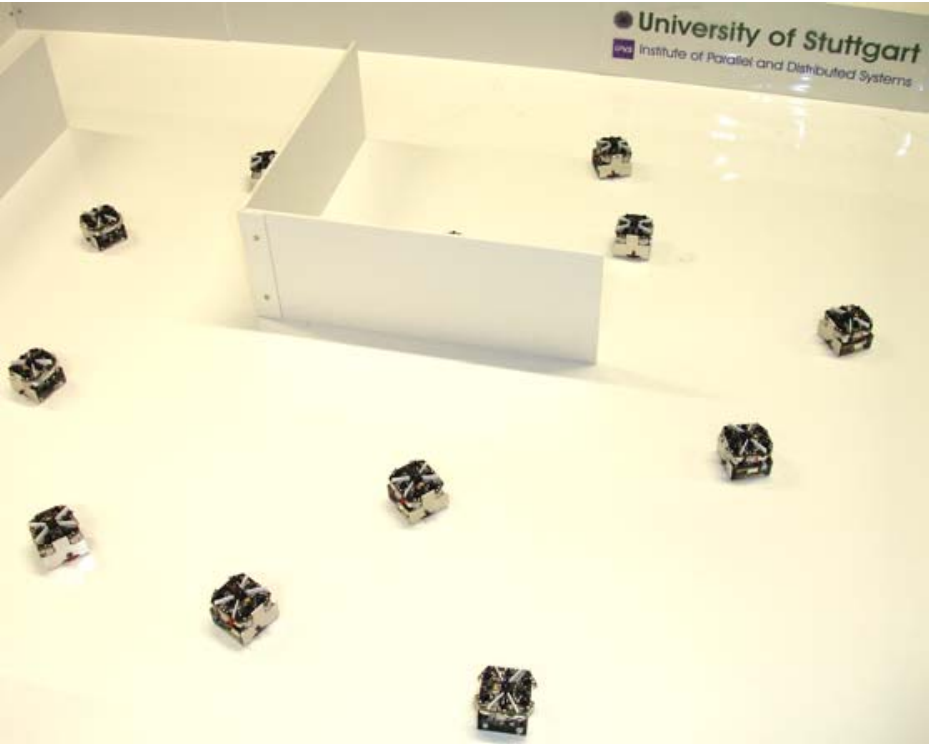}}\\
\subfigure[\label{fig:experiments3}]{\includegraphics[width=.42\textwidth]{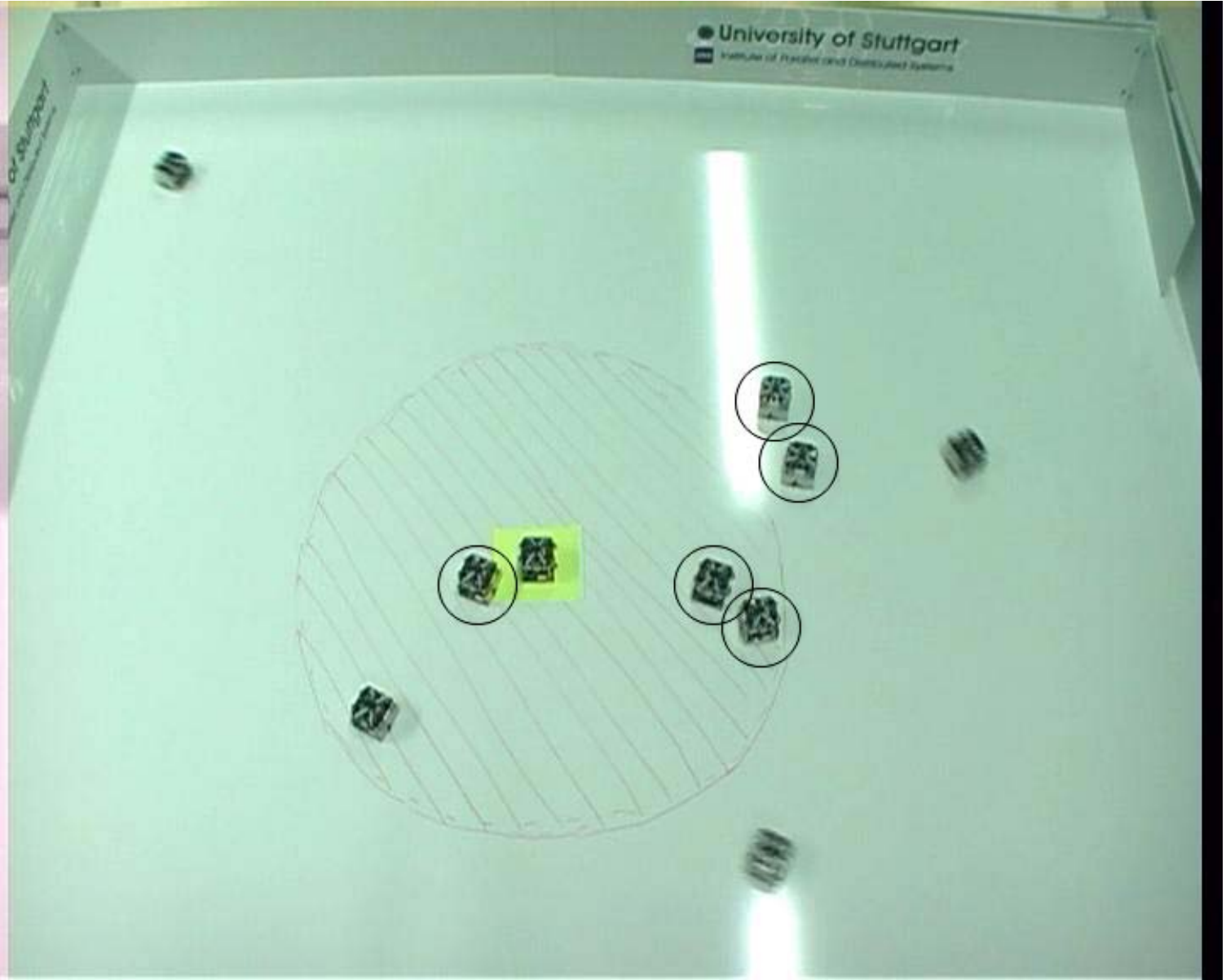}}~
\subfigure[\label{fig:experiments4}]{\includegraphics[width=.58\textwidth]{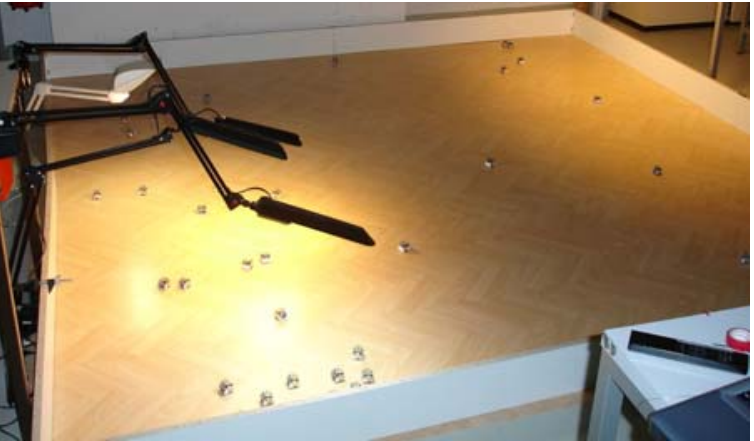}}\\
\subfigure[\label{fig:experiments5}]{\includegraphics[width=.495\textwidth]{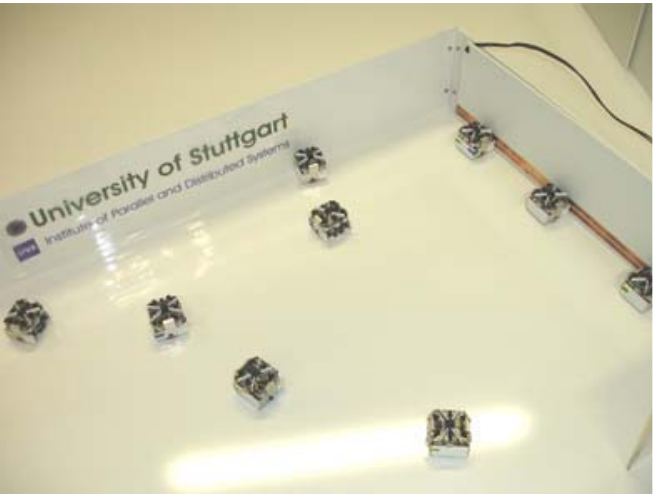}}~
\subfigure[\label{fig:experiments6}]{\includegraphics[width=.49\textwidth]{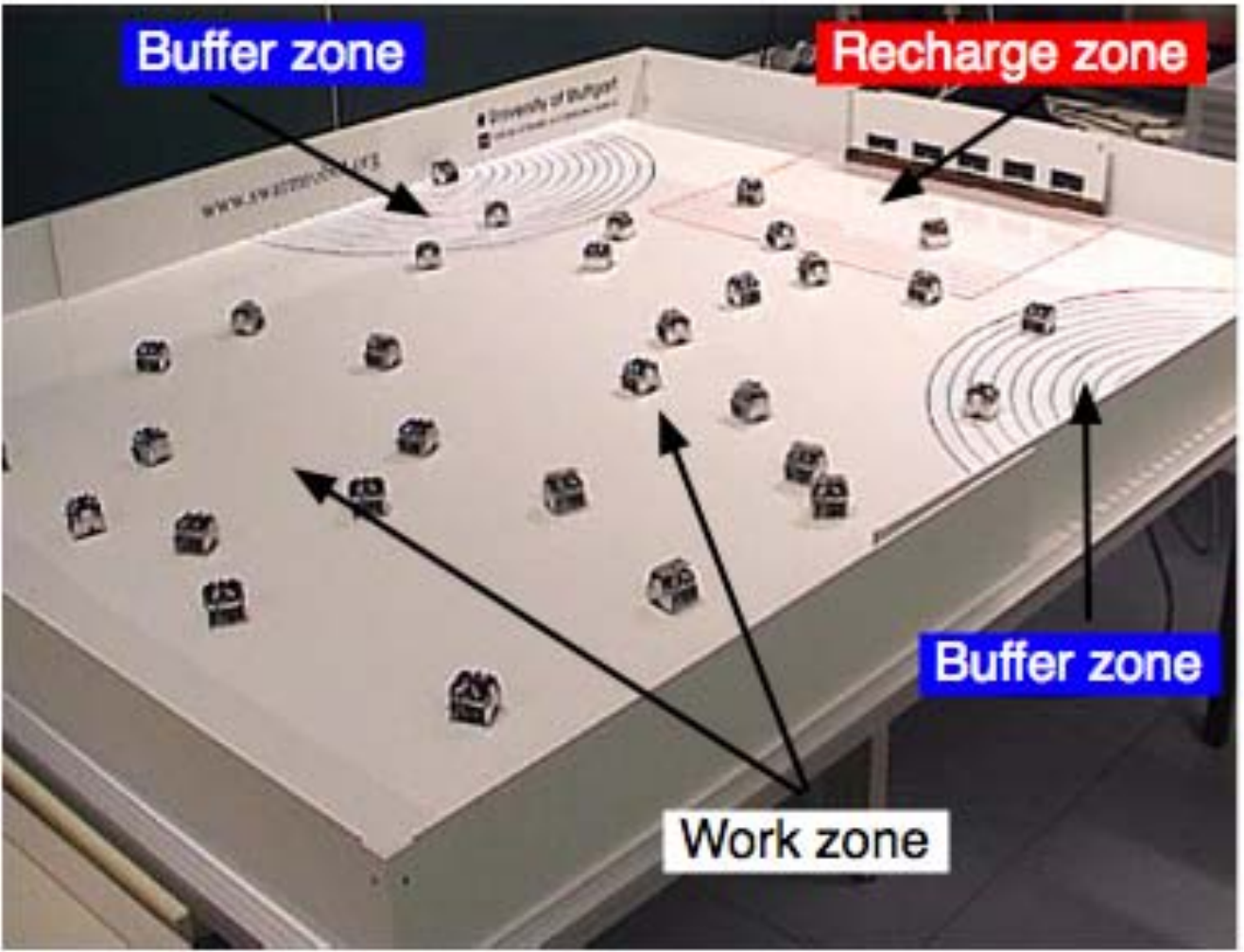}}
\caption{\small \textbf{(a)} Experiments with collective perception~\cite{Kornienko_S05a}; \textbf{(b)} Experiments swarm communications~\cite{KornienkoS05d}, \cite{2011arXiv1109.4221K}; \textbf{(c)} Experiments with tech-inspired aggregation~\cite{arXiv:1110.5183v1}; \textbf{(d)} Experiments with bio-inspired aggregation~\cite{Kornienko_S06}; \textbf{(e)} Experiments with docking and recharging~\cite{Kornienko_S06b}; \textbf{(f)} Experiments with bio-inspired collective energy management~\cite{Kernbach09Nep}.
\label{fig:experiments}}
\end{figure}

Currently there are a few swarms with Jasmine robots at different institutions. In our lab, we focused primarily on exploring various ways of deriving local rules, creating artificial self-organization and different kind of emergent phenomena. There are four main approaches in developing such local rules: top-down, bottom-up, evolutionary and bio-inspired, see e.g.~\cite{Crespi08}, \cite{Isakowitz98}, \cite{Kornienko_S04a}, \cite{McFarland86}, \cite{Kernbach08}, \cite{Pizka04} and others. The same problem can be considered in light of these four approaches: in several experiments we explored and compared their output and performance (e.g. bio-inspired~\cite{Kernbach09Nep} and tech-inspired ~\cite{Kornienko_S06b} approaches). Especial attention has been paid to basic problems such as collective energy management, swarm communication, collective and perception and collective awareness. Several images of the performed experiments are shown in Fig.~\ref{fig:experiments}. Many our projects, such as~\cite{I-Swarm}, \cite{golem}, \cite{ANGELS}, \cite{symbrion}, \cite{evobody}, \cite{replicator}, \cite{cocoro} address different aspects of the performed research.\\

\textbf{Bottom-up approach.} The local rules are first programmed into each agent and then cyclically evaluated and redesigned, see rule-based programming~\cite{Roma93}, refining sequential program into concurrent ones~\cite{Back91}, formal definition of cooperation and coordination \cite{Back88}. There are several variations of this technique: the application of optimization \cite{Chen04} or probabilistic \cite{Pradier05} approaches for finding optimal rules, geometrical \cite{Fu05} and functional \cite{Warraich05} considerations, the exploration of different aspects of embodiment: properties of communication \cite{Caselles05}, \cite{Mletzko06}, \cite{Geider06}, power management \cite{Jebens06}, \cite{Attarzadeh06} and sensing \cite{Zetterstrom06}.

\textbf{Top-down approach.} Macroscopic behavior is formally described, see for example grammatical and semantical structures \cite{Muscholl01}. By using a formal transformation, this high-level description can be converted to low-level programs in each collective agent. The top-down approach works well in different fields of nonlinear dynamics~\cite{Haken84}, and in the application of analytical approaches for controlling collective behavior~\cite{Levi99},~\cite{Kornienko_S99}, collective decision-making~\cite{Kornienko_OS01} and similar problems. Several optimization approaches can be used to perform top-down derivation of local rules for industrial environments \cite{Kornienko_OS01}, \cite{Kornienko_S03}, \cite{Kornienko_S03A}, \cite{Kornienko_S03C}, \cite{Kornienko_S04}. In robotic systems, top-down approaches have been applied to cooperative actuation \cite{Jimenez05}, \cite{Mletzko06a}, creation of desired behavioral patterns \cite{Prieto06}, and self-assembly processes \cite{Kabir08}, \cite{Ruth09}.

\textbf{Evolutionary approach.} The search space of a collective system contains a desired solution, that is, local rules, which are able to create a desired behavior \cite{Nolfi2000Evolutionary-Ro}. This desired solution is described by the fitness function \cite{branke2008edoeb}. Applying the principles of computational evolution, the developer can find the required rules. Such an approach has been applied to foraging problems, for example \cite{Koza92}, simple behavioral primitives  \cite{Koenig07_2}, genetic frameworks \cite{Nagarathinam07}, and for evolving morphology, controllers, behavior, or strategies.

\textbf{Bio-inspired approach.} Observations from social insects, animals, micro-organisms, or even humans are transferred to technical systems~\cite{Floreano08}. The number of bio-inspired works in the domain of collective systems, especially swarms, is very large. To give some examples, there are several attempts towards hormone-based regulation~\cite{Speidel08}, artificial sexual reproduction~\cite{Schwarzer08}, aggregation strategies inspired by bees \cite{Kornienko_S06}, bio-inspired decision-making \cite{Habe07} and foraging \cite{Kancheva07}, \cite{Kernbach09Nep}.

\textbf{Collective perception.} This problem was addressed in several works based on recognition of colors~\cite{Zetterstrom06} and geometries~\cite{Pradier05}, \cite{Kornienko_S05a}, \cite{Kornienko_S05d}. They are related to recognition of such objects that are larger than the robot itself. For that, all robots surround an object and scan the corresponding object's surfaces (see Fig.~\ref{fig:experiments1}). The scan data provide information about surface's geometry and allow classifying the type of surface (flat, concave, convex; size of surface and so on). The classification data are exchanged between robots and matched with the distributed model of the object. However for a particular robot is important not only to recognize an object, but also to know its own position in relation to this object. This positional context cannot be obtained from the sensor data of individual robot. The idea here is that during local communication, all robots know their neighbors, and this "embodied" information can be used for estimating a position. When sequences of connected particular observations are matched with models (e.g. the model A-B-C-A-C-D and the connected particular observation A-B), these connected observations can be located in the model. In this way the robots can collectively estimate their own spatial context (see more in~\cite{Pradier05}). As demonstrated by experiments, even for uncomplete observations, this approach allows deriving the positions.

\textbf{Spatial information processing.} These experiments are focused on spatial information processing, such as distance/area measurement over a swarm, collective calculation of the swarm's center of gravity~\cite{Fu05}, spatial decision making~\cite{Habe07} and others. These works are related in many aspects to cooperative actuation~\cite{Jimenez05}, embodiment issues \cite{Caselles05}, \cite{Mletzko06a}, cooperation and functional self-organization \cite{Warraich05}. To perform these operations, the robots have to create the dynamic communication network so that a part of them is continuously contained within the communication radius of each other. In this way they build a collective information system, which we call a "communication street".

\textbf{Swarm communication.} Communication approaches are integrated into other algorithms and into the software library of collective behavior. Several of these works are published in relation to IR-based~\cite{KornienkoS05d}, ZigBee~\cite{Beurer08}, \cite{Nagarathinam07} approaches, to using context-based communication~\cite{Kornienko_S05b} and to high-level communication procedures \cite{2011arXiv1109.4221K}, \cite{arXiv:1110.5183v1}, \cite{Kornienko_S06b}. Several aspects of these works are related to maintain the dynamical communication networks, optimization of energy and using the context provided by the amplitude and direction of IR signals. In this way they not only support an optimal distance for global information propagation, but also can collectively perform different spatial computations (distances, areas, centers of gravity and others).

\textbf{Feedback-based aggregation.} Aggregation was considered as a typical problem (task) for swarms and was solved via bio-inspired~\cite{Kornienko_S06}, \cite{SchmicklGet09}, \cite{Habe07}, \cite{Kancheva07}, bottom-up \cite{Mletzko06a}, \cite{Zetterstrom06} \cite{Prieto06}, top-down \cite{Kernbach08}, \cite{Kornienko_S05e} and evolutionary \cite{Koenig07}, \cite{Koenig07_2} methods. The example of aggregation was also used to compare scalability, performance and efficiency of different approaches.

\textbf{Energy-related aspects.} Energy aspects are investigated from different points of view: docking and low-level power management \cite{Jebens06}, \cite{Attarzadeh06}, bio-inspired \cite{Kernbach09Nep} and tech-inspired~\cite{Kornienko_S06b} behavioral strategies and energy homeostasis. We performed also several long-term experiments to explore self-sustainability of robot swarm and autonomous energy management.

\textbf{Self-assembling and multi-cellularity.}  These issues are primarily related to the projects \cite{symbrion}, \cite{replicator}, where we used another hardware platform~\cite{Kernbach09Platform}, \cite{Kernbach08_2}, \cite{Kernbach08Permis}, \cite{kernbach09adaptive} or the in edited book \cite{Levi10}. However, initial experiments as well as swarm-related topics in these projets are still explored with Jasmine robots~\cite{Kornienko_S07}.

\textbf{Collective awareness.} These works are relatively new and related primarily to the project~\cite{cocoro}. Here we explore mechanisms, which create awareness about a global state of the swarm on a local level without using any centralized mechanisms.

\section{Conclusion}

In this paper we presented an open-hardware project that targets a development of a cheap and reliable microrobot as well as SDK for large-scale swarms. We demonstrated the concept of a swarm embodiment, which underlies the development of robot platform Jasmine. Its capabilities and features are briefly described and illustrated by examples. Software part and simulator are briefly described.  Generally, we have shown that despite the limited hardware capabilities of microrobots, the specific construction of hardware and software parts make feasible many advanced collective properties.

The performed experiments between 2005 and 2010 are briefly listed in categories related to research topics explored in the group. However, a systematic consideration of all performed experiments and their evaluation in terms of embodiment, performance, scalability and other features still remain open and represents future works.

\section{Acknowledgment}

The \emph{swarmrobot.org} project was started in a close collaboration with Marc Szymanski who worked at those time at the University of Karlsruhe. His enthusiasm and energy essentially contributed the success of this project.

Author and the team are supported by the followings grants: EU-IST FET
project "I-SWARM", grant agreement no. 507006; EU-NMP
project "Golem", grant agreement no. 033211; EU-IST FET
project "SYMBRION", grant agreement no. 216342; EU-ICT
project "REPLICATOR", grant agreement no. 216240; EU-ICT
project "EvoBody", grant agreement no. 258334; EU-ICT FET
project "Angels", grant agreement no. 231845; EU-ICT
project "CoCoRo", grant agreement no. 270382.

\small

\end{document}